\documentclass[]{article}
\usepackage{spconf,amsmath,graphicx,hyperref}
\usepackage{booktabs} 
\usepackage{multirow}
\usepackage{array}
\usepackage{amsmath}
\usepackage{amssymb}
\usepackage{xcolor}
\usepackage{subcaption}

\title{LSTC-MDA: A Unified Framework for Long-Short Term Temporal Convolution and Mixed Data Augmentation in Skeleton-Based Action Recognition}
\name{Feng Ding, Haisheng Fu, Soroush Oraki, Jie Liang$^{*}$\thanks{*: Corresponding author.\\
\hspace*{\parindent} This paper is supported by the Natural Sciences and Engineering Research Council of Canada (NSERC) under grant RGPIN-2020-04525.}}
\address{School of Engineering Science,\\
	Simon Fraser University, BC, Canada\\
	\{feng\_ding, haisheng\_fu, soroush\_oraki, jie\_liang\}@sfu.ca}
\begin{document}
%
\maketitle
\begin{abstract}
Skeleton-based action recognition faces two longstanding challenges: the scarcity of labeled training samples and difficulty modeling short- and long-range temporal dependencies. To address these issues, we propose a unified framework, \emph{LSTC-MDA}, which simultaneously improves temporal modeling and data diversity. We introduce a novel Long-Short Term Temporal Convolution (LSTC) module with parallel short- and long-term branches, these two feature branches are then aligned and fused adaptively using learned similarity weights to preserve critical long-range cues lost by conventional stride-2 temporal convolutions. We also extend Joint Mixing Data Augmentation (JMDA) with an \emph{Additive Mixup} at the input level, diversifying training samples and restricting mixup operations to the same camera view to avoid distribution shifts. Ablation studies confirm each component contributes. LSTC-MDA achieves state-of-the-art results: \textbf{94.1\%} and \textbf{97.5\%} on NTU 60 (X-Sub and X-View), \textbf{90.4\%} and \textbf{92.0\%} on NTU 120 (X-Sub and X-Set),\textbf{97.2\%} on NW-UCLA. Code: \href{https://github.com/xiaobaoxia/LSTC-MDA}{https://github.com/xiaobaoxia/LSTC-MDA}.
\end{abstract}

\begin{figure*}[th]
  \centering
  \includegraphics[page=2,width=\linewidth]{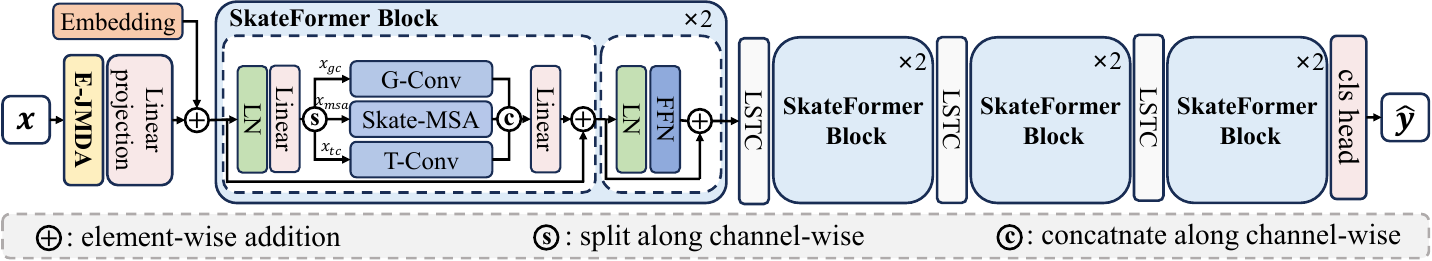}
  \caption{Pipeline of the proposed method. Given input $x$, Enhanced JMDA (E-JMDA) is applied to increase sample diversity. The output is then passed through a linear projection and added with positional embedding. SkateFormer blocks and the LSTC module are used to extract classification features, which are finally fed into a classification head to get the output label $\hat{y}$.}
  \label{fig:pipeline}
\end{figure*}
\section{Introduction}
\label{sec:intro}
Skeleton-based action recognition is attractive for its low computational cost and privacy benefits~\cite{liang2024skeleton}. Unlike RGB‐based methods, skeleton approaches use high-level joint and motion representations. Although RGB-based methods achieve high performance with CNNs, pretrained weights and optical flow~\cite{simonyan2014two}, they remain vulnerable to clutter, occlusions, and viewpoint variations. In contrast, skeleton representations abstract away low‐level visual details, producing compact and factor‐invariant features that are robust and computationally efficient, enabling more consistent and reliable performance in complex, unconstrained environments~\cite{sun2018pwc}.

Deep learning enhances skeleton recognition via GCNs~\cite{kipf2016semi} and Transformers~\cite{vaswani2017attention}. GCNs, e.g., ST-GCN~\cite{yan2018spatial} and CTR‐GCN~\cite{chen2021channel} model spatial and temporal relationships but often face challenges in capturing long‐range temporal dependencies. In contrast, Transformer‐based models~\cite{zhou2022hypergraph,do2024skateformer} use self‐attention to globally capture correlations among joints, offering more flexibility. 
However, these methods require large labeled datasets. GCN- and Transformer-based models struggle with physical constraints and overfit on limited data. Moreover, many approaches use suboptimal temporal downsampling. In particular, most methods~\cite{zhou2022hypergraph,do2024skateformer} perform temporal downsampling using a convolution with kernel size $7$ and stride $2$, capturing only short-term dependencies and misses critical long-range correlations. For instance, distinguishing ``\textit{put on}'' vs. ``\textit{take off}'' a shoe requires long-term context beyond short-term patterns.
Moreover, Skeleton-based methods also have smaller datasets than image/video approaches. Existing skeleton benchmarks (e.g., NTU RGB+D~\cite{liu2019ntu}) suffer from occlusions and pose estimation noise. 
JMDA~\cite{xiang2025joint} mitigates this issue via data augmentation but ignores cross-view inconsistencies, producing unrealistic poses.

To overcome these issues, we propose a Long–Short Term Temporal Convolution (LSTC) to replace standard temporal downsampling. LSTC captures short- and long-range dependencies via short- and long-term branches. Their outputs are linearly aligned and adaptively fused into the downsampled representation. Meanwhile, we extend JMDA with input-level additive Mixup and view-consistent augmentation within each camera to avoid unrealistic samples. LSTC-MDA achieves SOTA on NTU RGB+D 60/120 and NW-UCLA with minimal extra computation. We further provide comprehensive analyses to validate the complementary effects of each component and offer insight for future work.

\section{METHODOLOGY}
\begin{figure}[t]
  \centering
  \includegraphics[page=4,width=\linewidth]{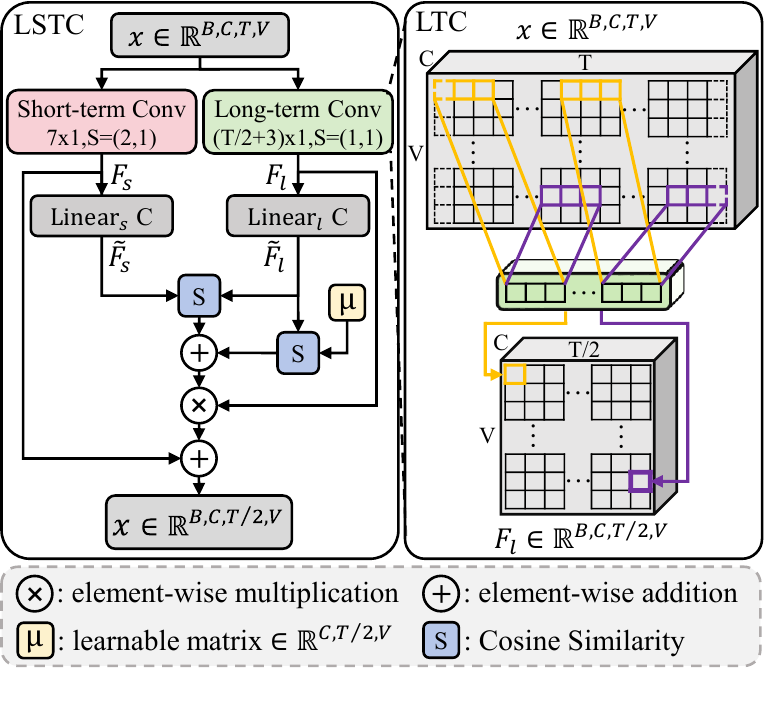}
  \caption{
  Long-Short Term Temporal Convolution (LSTC) in LSTC-MDA.
  \textbf{Left:} Two-branch design: short-term branch extracts $F_{s}$ using a $(7,1)$ convolution with stride 2; long-term branch computes $F_{l}$ via a specialized long-term convolution. Both features are linearly aligned to $\tilde{F}_{s}$ and $\tilde{F}_{l}$, and their similarity, combined with an auxiliary similarity between $\mu$ and $\tilde{F}_{l}$, is used as a weight to fuse $F_{s}$ and $F_{l}$. 
  \textbf{Right:} Long-term convolution with $(T/2 + 3)$ kernel that only processes the first and last three positions, skipping intermediate elements.
  }
  \label{fig:LSTC}
\end{figure}
\subsection{Preliminaries}
A skeleton sequence is a tensor $\mathbf{X} \in \mathbb{R}^{C \times T \times V \times M}$, where $C$ is the number of feature channels (e.g., 3D joint coordinates), $T$ is the number of frames, $V$ is the number of joints per body, and $M$ is the maximum number of bodies. After preprocessing and data augmentation, we merge the body and joint dimensions: $V \leftarrow V \times M$, yielding $\mathbf{X} \in \mathbb{R}^{C \times T \times V}$.

We adopt SkateFormer~\cite{do2024skateformer} as our backbone. The input is linearly projected and combined with SkateEmbedding~\cite{do2024skateformer}. Our modifications are applied on top of this pipeline. Between SkateFormer stages, we replace the standard downsampling with LSTC module,
resulting in three LSTC layers among four stages. After the final stage, the feature tensor has shape $\mathbb{R}^{2C \times \frac{T}{8} \times V}$. A global pooling layer followed by a linear classification head is applied
to produce the output $\hat{y}$.
\begin{table*}[!htbp]
\centering
\setlength{\tabcolsep}{4pt} 
\renewcommand{\arraystretch}{1.1} 
\small
\begin{tabular}{c|c|c|c|c|c|c|c|c|c|c|c|c|c|c}
\hline
&  &
\multicolumn{6}{c|}{\textbf{NTU RGB+D (\%)}} &
\multicolumn{6}{c|}{\textbf{NTU RGB+D 120 (\%)}}\\
\cline{3-14}
\textbf{Methods} & \textbf{Venue} & \multicolumn{3}{c|}{\textbf{X-Sub60}}  & \multicolumn{3}{c|}{\textbf{X-View60}} & \multicolumn{3}{c|}{\textbf{X-Sub120}} & \multicolumn{3}{c|}{\textbf{X-Set120}} & 
\textbf{NW-UCLA} (\%) \\
\cline{3-14}
& & $E_1$ & $E_2$ & $E_4$ & $E_1$ & $E_2$ & $E_4$ & $E_1$ & $E_2$ & $E_4$ & $E_1$ & $E_2$ & $E_4$ \\
\hline


HD-GCN~\cite{lee2023hierarchically} & ICCV2023 & 90.6 & 92.4 & 93.0 & 95.7 & 96.6 & 97.0 & 85.7 & 89.1 & 89.6 & 87.3 & 90.6 & 91.2 & 96.9 \\
STC-Net~\cite{lee2023leveraging} & ICCV2023 & - & 92.5 & 93.0 & - & 96.7 & 97.1 & - & 89.3 & 89.9 & - & 90.7 & 91.3 & 97.2 \\
ProtoGCN~\cite{liu2025revealing} & CVPR2025 & 91.6 & 93.0 & 93.5 & 96.3 & 97.2 & 97.5 & 85.5 & 89.8 & 90.4 & 88.4 & 91.2 & 91.9 & - \\

BlockGCN~\cite{zhou2024blockgcn} & CVPR2024 & - & - & 90.9 & - & - & 95.4 & - & - & 86.9 & - & - & 88.2 & 95.5 \\
LA-GCN~\cite{xu2025language} & TMM2025 & - & 92.3 & 93.0 & - & 96.6 & 97.1 & 86.5 & 89.7 & 89.9 & 88.0 & 90.9 & 91.3 & 96.8 \\
Hyperformer~\cite{zhou2022hypergraph} & Arxiv2022 & 90.7 & - & 92.9 & 95.1 & - & 96.5 & 86.6 & - & 89.9 & 88.0 & - & 91.3 & 96.7 \\
\hline
SkateFormer~\cite{do2024skateformer} & ECCV2024 & {92.6} & {93.0} & {93.5} & {97.0} & \textcolor{red}{97.4} & \textcolor{red}{97.8} & {87.7} & {89.4} & {89.8} & {89.3} & {91.0} & {91.4} & \textcolor{red}{98.3} \\
SkateFormer-R & ECCV2024 & \textcolor{blue}{92.4} & {93.0} & \textcolor{blue}{93.3} & \textcolor{blue}{96.8} & \textcolor{blue}{97.2} & \textcolor{blue}{97.3} & {87.7} & {89.4} & \textcolor{blue}{89.7} & {89.3} & \textcolor{blue}{90.7} & \textcolor{blue}{91.1} & \textcolor{blue}{95.9} \\
LSTC-MDA (Ours) &  & \textcolor{red}{\textbf{93.4}} & \textcolor{red}{\textbf{93.9}} & \textcolor{red}{\textbf{94.1}} & \textcolor{red}{\textbf{97.0}} & \textbf{97.3} & \textbf{97.5} & \textcolor{red}{\textbf{89.0}} & \textcolor{red}{\textbf{90.2}} & \textcolor{red}{\textbf{90.4}} & \textcolor{red}{\textbf{90.7}} & \textcolor{red}{\textbf{91.7}} & \textcolor{red}{\textbf{92.0}} & \textbf{97.2} \\
\hline
\end{tabular}

\caption{Comparison on NTU RGB+D and NW-UCLA datasets under various evaluation settings.
Numbers in \textcolor{red}{red} indicate the best performance in each setting.
Numbers in \textcolor{blue}{blue} denotes reproduced results below the original paper~\cite{do2024skateformer}}
\label{tab:result}
\end{table*}
\subsection{Long-Short term Temporal Convolution}
The temporal convolution (T-Conv) used in SkateFormer is just a simple convolution layer with stide is 2 and kernel size is (7,1), which primarily captures short range neighboring pattern, but missing long-term dependencies. To address this, we propose the Long-Short Term Temporal Convolution (LSTC), shown in Fig.~\ref{fig:LSTC}. LSTC consists of a short-term and a long-term branch. The short-term feature $F_{s}\in \mathbb{R}^{C \times T/2 \times V}$ and long-term feature $F_{l}\in \mathbb{R}^{C \times T/2 \times V}$ are extracted using separate convolution layers. The features are aligned via linear projections for similarity computation. Cosine similarities $S_{sl}$ are computed between the aligned features $\tilde{F}_{s}$ and $\tilde{F}_{l}$, and combine it with an auxiliary similarity $S_{\mu l}$ between a learnable matrix $\mu\in \mathbb{R}^{C \times T/2 \times V}$~\cite{xiao2023lstfe} and $\tilde{F}_{l}$. These similarities generate adaptive weights to fuse $F_{s}$ and $F_{l}$:
\begin{equation}
x = F_{s}+(S_{sl}+S_{\mu l})\times F_{l}.\label{eqn:5}
\end{equation}
The short-term branch applies a $7\times1$ convolution with stride $(2,1)$ to capture local patterns.
The long-term branch adopts a sparse convolution with a large receptive field, as illustrated in Fig.~\ref{fig:LSTC}. Where only the first and last three weights are learnable and all intermediate weights are zero, resulting in a small number of learnable parameters with a shape of $(C, C, 6, 1)$, introducing minimal additional computation. This design ensures that the convolution attends only to the first 3 and last 3 temporal positions, ignoring the intermediate frames. Formally, the kernel $w$ is non-zero only at indices $I={0,1,2,\tfrac{T}{2},\tfrac{T}{2}+1,\tfrac{T}{2}+2}$, as in Eq.~\ref{eqn:6}.

\begin{equation}
y[n] = \sum_{c}^{C}\sum_{i\in I} w_{c,i} \cdot x[n + i],\label{eqn:6}
\end{equation}
where $c$ and $C$ represent the current and total channel. 

\subsection{Enhanced JMDA}
JMDA~\cite{xiang2025joint} first employs a feature alignment method to extend the input skeletons that contain only a single human body or limited frames—to match the maximum number of bodies and frames in the dataset. In this way, all data have same number of human and frames, with no empty space, which is beneficial for data augmentation. It has two modules: SpatialMix and TemporalMix, which augment spatial and temporal dimensions. We extend JMDA with an \emph{Additive Mixup} at the input level, diversifying training samples and restricting mixup to the same camera view to avoid distribution shifts.

\textbf{AdditiveMix.}
We adopt a mixup strategy based on linear interpolation:
\begin{equation}
\begin{gathered}
\lambda_a \sim \mathrm{Beta}(2, 2) \\
x_m = \lambda_a x_i + (1 - \lambda_a) x_j \quad
y_m = \lambda_a y_i + (1 - \lambda_a) y_j.
\end{gathered}\label{eqn:11}
\end{equation}
Here, a mixing coefficient $\lambda_a$ is sampled from a $Beta$ distribution as in mixup~\cite{zhang2017mixup}, and samples $x_i$ and $x_j$ are linearly combined to generate new training data.

\textbf{View-Consistent Group-Wise Mixup.} 
To ensure effective data augmentation under multi-view scenarios, we apply view-consistent mixup within each camera group for the NTU RGB+D 60 (cross-view), NTU RGB+D 120 (cross-set), and NW-UCLA benchmarks. Existing augmentation methods typically neglect view inconsistency in multi-view scenarios, even though these benchmarks involve training and testing sets captured from different cameras with varying viewpoints. Performing mixup across disparate views can result in unrealistic samples with artificial intermediate angles that deviate from the true data distribution. Such inconsistencies may reduce augmentation effectiveness and even degrade model performance. By restricting mixup within-camera groups, we maintain view consistency and improve the generalization of the augmented samples.

Finally, the three augmentation strategies TemporalMix, SpatialMix, and AdditiveMix are applied jointly to enhance training diversity. Each is applied independently with a probability of 50\%, resulting in up to 10 possible variants per sample, including $x$, $T(x)$, $S(x)$, $A(x)$, and their permutations ($S(T(x))$, $A(T(S(x)))$), where $T$, $S$, and $A$ denote the respective augmentation methods. Compared to the original JMDA, which produces only five variants using TemporalMix and SpatialMix, our enhanced strategy significantly increases sample diversity with minimal additional computational cost.

\section{Experiments}

\subsection{Implementation Details}
We evaluate our method on NTU RGB+D~\cite{liu2019ntu} and NW-UCLA~\cite{wang2014cross}, following SkateFormer~\cite{do2024skateformer}.
Experiments use PyTorch on a single RTX 4090 GPU. We reimplemented SkateFormer (SkateFormer-R in Table~\ref{tab:result}) as a baseline with slightly lower accuracy than reported. The slight discrepancy may be due to the hyper-parameter details not fully specified in the original paper, along with differences in CUDA and PyTorch versions. All proposed enhancements are built on this reproduced baseline.

Models were trained for 500 epochs with batch size 128. Learning-rate warmed up linearly from $1\times10^{-7}$ to $1\times10^{-3}$ over 25 epochs, then followed a cosine schedule. We used AdamW ($\beta_{1}=0.9,\ \beta_{2}=0.999$, weight decay 0.1), gradient clipping (max norm 1), random seed (1), and KL-divergence loss. All other architectural settings (e.g., hidden dimensions, embedding schemes) follow SkateFormer configuration.

\subsection{Comparison with the State of the Art}
We compare the performance of different methods across three modality‐ensemble settings ($E_1$, $E2$, and $E4$): (i) $E1-$joint modality only; (ii) $E2-$joint and bone modalities; and (iii) $E4-$ joint, bone, joint motion, and bone motion modalities. Following prior work, we train separate models for each modality and ensemble their outputs. Table~\ref{tab:result} shows results on NTU RGB+D, NTU RGB+D 120, and NW-UCLA.

As shown in Table~\ref{tab:result}, LSTC‐MDA, outperforms prior SOTA methods across most benchmarks. While our results for NTU RGB+D X‐View60 $E2$ and $E4$ settings are slightly below the original SkateFormer, our method consistently surpasses the baseline, demonstrating the effectiveness of our improvements. 
Notably, under the $E2$ setting (using only joint and bone modalities), our method exceeds the $E4$ performance of several SOTA methods that use all four modalities. This indicates that our approach achieves competitive performance with fewer modalities and at a lower computational cost, making multi‐modality ensemble approaches more practical and efficient.
Compared with baseline, LSTC-MDA shows notable gains on fine-grained actions such as ``take off **" and ``put on **", highlight the importance of modeling local and global temporal dependencies when dealing with fine and similar actions in skeleton-based recognition.
\subsection{Ablation Study}
We conduct a series of ablation studies to examine the impact of each module in LSTC-MDA.
\begin{table}[htb!]
\centering
\setlength{\tabcolsep}{4pt} 
\renewcommand{\arraystretch}{1.1} 
\begin{tabular}{l|c}
\hline
\multicolumn{1}{c|}{Models}
&X-Sub60 ($E_1$)  \\
\hline
Skateformer-R (baseline)    & 92.42\%             \\
 + JMDA                                          & 93.19\%   (+0.77)   \\
+ JMDA + add-mix                        & 93.26\%   (+0.84)   \\
+ LSTC                  & 92.58\%   (+0.16)   \\
+ LSTC + JMDA           & 93.24\%   (+0.82)   \\
\hline
+ LSTC + JMDA + add-mix & 93.36\% (+0.94)\\
\hline
\end{tabular}
\caption{Ablation study of LSTC-MDA components on NTU RGB+D 60 Cross-Subject ($E_1$).
``SkateFormer-R” is the reproduced baseline; ``+JMDA”, ``+add-mix”, and ``+LSTC” indicate the incremental inclusion of JMDA, additive mixup, and LSTC, respectively.
}
\label{tab:ablation_components}
\end{table}

\textbf{Effectiveness of LSTC-MDA Components.}  
We examine the contributions of the data augmentation strategies and LSTC. As shown in Table~\ref{tab:ablation_components}, we use SkateFormer-R on X-Sub60 ($E_1$) as baseline. Each component individually improves performance. The complete LSTC-MDA achieves 93.36\%, showing complementary gains.

\begin{table}[ht]
\centering
\setlength{\tabcolsep}{4pt} 
\renewcommand{\arraystretch}{1.1} 
\begin{tabular}{l|c|c|c}
\hline
\multicolumn{1}{c|}{\multirow{2}{*}{Kernel Settings}} & X-Sub60 & NW-UCLA & \multicolumn{1}{c}{\multirow{2}{*}{$\Delta$Params}}\\
\cline{2-3}
&\multicolumn{2}{c|}{$E_1$}&\\
\hline
first 4\&last 4               & 93.45\% & 95.04\%& 0.942M\\
5 frames uniform & 93.32\% & 95.47\%& 0.678M\\
every other frame& 93.18\% & 95.04\%& 1.030M\\
\hline
first 3\&last 3 (ours)       & 93.36\% & 95.47\%& 0.766M\\
\hline
\end{tabular}
\caption{Accuracy on NTU RGB+D 60 Cross-Set ($E_1$) and NW-UCLA ($E_1$) for different LSTC kernels. $\Delta$Params shows parameter change vs. SkateFormer. “First 3\&last 3” achieves the best trade-off between accuracy and complexity.}
\label{tab:ablation_kernelsize}
\end{table}
\textbf{Impact of Kernel Setting in LSTC.}  To study the effect of different receptive fields in the long-term branch, we evaluate different long-term convolution kernels in LSTC, result
 shown in Table~\ref{tab:ablation_kernelsize}. 
``First 4\&last 4" slightly improves X-Sub60 but drops NW-UCLA accuracy and increases parameters. 
On the other hand, the ``5 frames uniform" sampling achieves $0.04\%$ lower performance while reducing complexity. However, when more densely uniform sampled frames are used, we observe a decline in accuracy. This may be because the ``every other frame” kernel captures much denser temporal information than the short-term branch, resulting in a large discrepancy that hinders feature fusion due to a mismatch in temporal granularity between the two branches. ``First 3\&last 3" provides the best trade-off, capturing long-range dependencies while preserving fusion quality. This design effectively captures long-range dependencies while minimizing the gap with the short-term branch, whereas using excessive uniform sampling tends to widen this gap and degrade feature fusion.


\begin{table}[ht]
\centering
\setlength{\tabcolsep}{4pt} 
\renewcommand{\arraystretch}{1.1} 
\begin{tabular}{l|c|c}
\hline
\multicolumn{1}{c|}{Benchmarks}& LSTC-Aug w/o   &LSTC-Aug w (ours) \\
\hline
X-View60  ($E_1$)   & 96.87\% & 96.99\%       \\
X-Set120 ($E_1$)  & 90.56\% & 90.67\%\\
NW-UCLA  & 96.3\% & 97.2\%\\
\hline
\end{tabular}
\caption{Proposed model with or without the view-consistent group-wise mixup data augmentation.}
\label{tab:ablation_camera}
\end{table}

\textbf{Effect of Group-wise Mixup.}  
We evaluate view-consistent group-wise mixup on cross-view and cross-setup benchmarks. Specifically, we compare results with and without group-wise augmentation on X-View60 ($E_1$), X-Set120 ($E_1$) and NW-UCLA, where training and test samples are captured from different camera angles. Table~\ref{tab:ablation_camera} shows applying mixup in group-wise significantly improves performance by better aligning augmented samples with the test distribution.

\section{Conclusion}
We propose \emph{LSTC-MDA}, a framework that advances temporal modeling and data diversity for skeleton-based action recognition. The LSTC module integrates a short-term 7-feature convolution with a sparse long-range branch, fusing them via learned similarity weights to capture both fine-grained motions cues and global context. Building upon JMDA, we introduce input-level additive mixup and a view-consistent group-wise mixing, proposed here for the first time, to increase sample diversity while avoiding unrealistic cross-view artifacts. Extensive experiments on NTU RGB+D 60/120 and NW-UCLA demonstrate LSTC-MDA consistently outperforms prior SOTA with minimal additional computation. 
Looking ahead, replacing the fixed sparse kernel with a learnable temporal sampling or adaptive dilation could further discover the most informative time offsets, and integrate a dedicated hand/finger modeling to further enhance discriminability on fine-grained gestures and subtle manipulations.

\bibliographystyle{IEEEbib}
\bibliography{refs}

\end{document}